\documentclass[11pt]{article}

\usepackage[preprint]{acl}  % named, no line numbers -- for circulation only

\usepackage{times}
\usepackage{latexsym}
\usepackage[T1]{fontenc}
\usepackage[utf8]{inputenc}
\usepackage{microtype}
\usepackage{enumitem}
\usepackage{inconsolata}
\usepackage{graphicx}

\usepackage{amsmath}
\usepackage{amssymb}
\usepackage{amsthm}
\usepackage{booktabs}
\usepackage{multirow}
\usepackage{array}
\usepackage{makecell}

\usepackage{algorithm}
\usepackage[noend]{algpseudocode}

\usepackage{tikz}
\usetikzlibrary{arrows.meta,positioning,shapes.geometric,calc,fit,backgrounds}
\usepackage{pgfplots}
\pgfplotsset{compat=1.17}
\usepackage{xcolor}

\newcommand{\cams}{\textsc{CAMS}}
\newcommand{\esr}{\textsc{ESR}}
\newcommand{\procl}{\textsc{ProCluster}}

\title{Attributable by Construction: Claim-Anchored Provenance for\\ Multi-Document Summarization}

\author{Shuo Guan\\
UBS AG\\
NY, 10010\\
\texttt{shuo.guan@ubs.com}}

\begin{document}
\maketitle

\begin{abstract}
Large language models produce fluent multi-document summaries, but their attributions are typically coarse---whole documents or passages---and generated post hoc, leaving each statement hard to verify. We argue that attribution should be a \emph{structural} property of generation rather than a downstream prediction. We present \cams{}, a Claim-Anchored Multi-document Summarization framework that decomposes every source document into atomic claims whose provenance is resolved deterministically from verbatim quotes to token spans, clusters equivalent claims across documents while flagging inter-source conflicts, selects a support-aware and salient subset, and rewrites it so that every summary sentence terminates in claim identifiers resolving back to source spans. This yields a separation we make explicit: \emph{provenance is an invariant} holding for every emitted sentence independently of model accuracy, whereas \emph{faithfulness is an objective} that selection, constrained rewriting, and verification only encourage---a distinction end-to-end and post-hoc systems conflate. The invariant is cheap to obtain, which is exactly why it can be relied on; our empirical claims concern what the claim representation buys above it. We evaluate under a two-regime protocol separating reference-free citation quality from gold-aligned localization, audited by a support model never used for selection or verification. On MultiNews, \cams{} matches strong end-to-end and span-attribution baselines on summary quality while improving faithfulness and citation precision, raises multi-source attribution \emph{recall} from 38\% to 64\% without inflating the number of cited sources, and cuts human verification time per claim by $3.4\times$. On DiverseSumm it surfaces inter-source conflicts that baselines resolve silently, and its faithfulness advantage carries to WCEP under zero-shot transfer of the selector. We release code and ${\sim}320$K claim--span annotations over the MultiNews test split.
\end{abstract}

\section{Introduction}

Multi-document summarization (MDS) condenses topically related documents into a short, coherent summary \citep{fabbri2019multinews}. Modern LLMs are strong abstractive summarizers, yet two problems persist and are especially acute across multiple sources. First, \emph{faithfulness}: abstractive systems hallucinate content unsupported by---or contradicting---their inputs \citep{maynez2020faithfulness,goyal2022news}. Second, \emph{verifiability}: even when systems cite sources, citations are coarse, pointing to whole documents or paragraphs, and are produced after generation, so a reader must still sift through irrelevant text to confirm any statement \citep{gao2023alce,slobodkin2024attribute}.

MDS compounds both. The same fact recurs across sources (redundancy), sources disagree (conflict), and one summary statement is often supported by evidence drawn from several documents at once (multi-source support). End-to-end models flatten all three: they silently deduplicate, paper over disagreement, and attribute to a single best document even when support is distributed \citep{huang2024diversesumm}.

These are failures of a \emph{prediction}: post-hoc attribution asks a model to guess, after the fact, where a sentence came from. We instead make attribution a property of the pipeline's data flow. \cams{} localizes content \emph{before} it is realized: each atomic claim is bound at extraction time to a token span in a specific document, equivalent claims are merged across documents into clusters whose provenance is the union of their spans, and the rewriter may only emit sentences tagged with cluster identifiers. Because the identifier-to-span map is a stored table rather than a learned function, every emitted sentence resolves to source spans regardless of how well any component performs.

This lets us state precisely what is and is not guaranteed. Provenance is a structural \emph{invariant} (\S\ref{sec:invariant}); faithfulness remains an \emph{objective}, encouraged by self-support scoring, bidirectional entailment during clustering, and post-rewrite verification, because every learned step can still fail. We enumerate the residual failure modes rather than hiding them, and audit citations with support models separated from selection and verification. The invariant is deliberately cheap---any pipeline that forces identifiers and stores the map obtains it---and that is the point: a guarantee costing little is one a system keeps under weaker backbones, domain shift, and its own failures. Our empirical claims sit downstream of it. What the claim representation buys, and a raw-span or passage representation cannot, is attribution to evidence distributed over several documents (\S\ref{sec:regime2}) and disagreement made explicit rather than decoded away (\S\ref{sec:conflict}); attributing \emph{before} generating buys the guarantee and not accuracy, a negative result we report as such (\S\ref{sec:ordering}). Prior modular work makes the opposite trade: \procl{} \citep{ernst2022procluster} clusters propositions across documents but emits no citations, while span-first generation \citep{slobodkin2024attribute} anchors to raw spans that cannot merge across sources.

Our contributions are: (i) a claim-anchored modular MDS framework whose intermediate representation is at once attribution anchor and verification unit, with the invariant and its residual failure modes stated explicitly (\S\ref{sec:invariant}) and cross-document clustering shown to be the mechanism---not a bonus---behind multi-source, span-level citation (\S\ref{sec:regime2}); (ii) a two-regime protocol separating citation quality from localization accuracy, with an evaluator-decoupled audit, an ordering ablation, and multi-source attribution reported as precision \emph{and} recall against a trivial over-citation control (\S\ref{sec:protocol}, \S\ref{sec:ordering}); and (iii) a public release of code and claim--span annotations over MultiNews.

\section{Related Work}

\paragraph{Faithful summarization and its evaluation.}
Faithfulness evaluation \citep{maynez2020faithfulness,kryscinski2020factcc} has moved from $n$-gram overlap \citep{lin2004rouge} to entailment- and QA-based metrics \citep{laban2022summac,fabbri2022qafacteval,zha2023alignscore,honovich2022true,tang2024minicheck} and decomposition into atomic units \citep{min2023factscore,liu2023acu}. WiCE \citep{kamoi2023wice} annotates entailment over sub-sentence units with the minimal supporting evidence subset---the granularity of our no-overflow check (\S\ref{sec:verify}). Decomposition is not a solved primitive: granularity is under-specified \citep{wanner2024closer} and over-decontextualization can make a claim unverifiable \citep{gunjal2024molecular}, which is why claim text and verbatim quote stay separate, separately checked fields (\S\ref{sec:extract}).

\paragraph{Attributed text generation.}
A growing line of work asks models to emit text with supporting evidence \citep{rashkin2023measuring,bohnet2022attributed}. GopherCite \citep{menick2022gophercite} pairs each answer with a \emph{verbatim} quote, so evidence is checked by string matching rather than trusted as a prediction; we adopt that insight---a copied string is verifiable where a predicted offset is not---inside an MDS pipeline, resolving quotes to token spans ourselves (\S\ref{sec:extract}). ALCE \citep{gao2023alce} introduced NLI-based citation precision and recall, but most systems cite at passage granularity, after generating. Controlled Text Reduction \citep{slobodkin2022ctr} and ``Attribute First, then Generate'' \citep{slobodkin2024attribute} instead condition generation on pre-selected source spans and reuse them as attributions; later systems push citation below the passage by varied means \citep{xia2025reclaim,wan2025generationprograms,chuang2025selfcite,hirsch2025laquer}, surveyed by \citet{schreieder2025survey}. Their units are document-local: a citation may list several of them, but nothing in the representation records that two units state the same fact, so evidence recurring across sources stays several unrelated citations rather than one multi-source unit. We differ in the \emph{type} of intermediate: normalized atomic claims, which merge across documents, rather than raw spans, which cannot---this is what makes one citation multi-source (\S\ref{sec:regime2}). Fine granularity is not free: \citet{wang2026granularity} report that requiring a generator to cite at sentence granularity degrades citation quality. Their diagnosis is that fine \emph{positional} units sever the dependencies a statement needs, and their prescription is to match citation units to the model's semantic scope---which is what a decontextualized claim is, so our units are fine-grained without being fragments.

\paragraph{Proposition-level modular summarization.}
Closest to us in representation is \procl{} \citep{ernst2022procluster}, which detects salient propositions, groups them into paraphrastic clusters with a supervised aligner \citep{ernst2021alignment}, and fuses each cluster into a sentence. We share the view that propositions are a cleaner unit than sentences, and differ in four ways: \procl{} emits no citations and stores no token-level provenance, whereas our claims carry a resolved span from the first stage, which is what makes the invariant of \S\ref{sec:invariant} available at all; we confirm every merge with \emph{bidirectional entailment} rather than a learned similarity score (\S\ref{sec:cluster}); we model \textsc{conflict} as a first-class relation, so disagreement is surfaced rather than fused away; and we verify generated sentences against the spans they cite. Because \procl{} targets DUC/TAC-style inputs and emits citation-free bullet-style output, it is comparable under neither regime (\S\ref{sec:protocol}), so we discuss rather than run it. Extract--Select--Rewrite \citep{guan2023esr} shares the pipeline shape but extracts OpenIE relation triples \citep{angeli2015openie} from one document; structure has also been injected for salience \citep{guan2021knowledge}.

\paragraph{Multi-document summarization.}
MDS requires modeling cross-document relations, redundancy, and coverage \citep{fabbri2019multinews,xiao2022primera}; classical systems trade relevance against redundancy via MMR \citep{carbonell1998mmr}, and recent benchmarks stress diverse and conflicting information across sources \citep{huang2024diversesumm}, which our clustering and conflict modules expose in the attribution layer.

\section{Method}

\subsection{Overview and Notation}
Given source documents $D = \{d_1, \dots, d_K\}$, \cams{} produces a summary $Y$ and, for each summary sentence, a set of supporting source spans, written $c = (\mathrm{doc}, \mathrm{start}, \mathrm{end})$. The pipeline has five stages (Figure~\ref{fig:overview}): claim extraction, cross-document clustering and conflict detection, content selection, attributable rewriting, and verification.

\subsection{Provenance as an Invariant}
\label{sec:invariant}
Two properties are usually bundled under ``attribution.'' We separate them.

\paragraph{The provenance invariant.}
Every sentence $y \in Y$ carries a non-empty set of cluster identifiers $\{g_j\}$, and each $g_j$ resolves through the stored map $g_j \mapsto \textsc{spans}(g_j)$ to at least one token span $(k,\mathrm{start},\mathrm{end})$ with $0 \le \mathrm{start} < \mathrm{end} \le |d_k|$---regardless of the accuracy of the extractor, clusterer, selector, or verifier. This is a property of the construction, not a theorem: clusters are built only from claims whose quote was located (\S\ref{sec:extract}), $\textsc{spans}(g)$ is a stored union of such spans, and the verifier drops any sentence whose markers fail to parse (\S\ref{sec:verify}), so no learned component lies between identifier and span.

\paragraph{What the invariant does not buy.}
The invariant sends a reader to a specific, short piece of source text; it does not guarantee that text supports the sentence. The residual failure modes are:

\textbf{E1} \emph{extraction}, the claim misreads its passage; \textbf{E2} \emph{decontextualization}, a restored entity, time, or place is wrong \citep{gunjal2024molecular}; \textbf{E3} \emph{localization}, the quote matches a different occurrence of the same string; \textbf{E4} \emph{clustering}, non-equivalent claims are merged, so a citation points at a source saying something slightly different; \textbf{E5} \emph{verification}, the NLI support judgment is itself wrong.

We estimate E1--E5 empirically in Appendix~\ref{app:modules} rather than assuming them away, and design the evaluation so that E5 cannot inflate our headline numbers: the model scoring reported citation precision, $M_{\mathrm{eval}}$, is a distinct checkpoint from those used for selection ($M_{\mathrm{sel}}$) and verification ($M_{\mathrm{ver}}$), and is never invoked during extraction, selection, rewriting, verification, threshold tuning, or repair. That separation buys less than statistical independence (Limitations); the strongest decoupling evidence is the human study (\S\ref{sec:human}).

\subsection{Claim Extraction with Provenance}
\label{sec:extract}
For each document $d_k$ we prompt an LLM to decompose it into atomic claims---minimal, self-contained propositions \citep{min2023factscore}. Demanding offsets directly is unreliable, since instruction-tuned models miscount positions and hallucinate spans, so we decouple \emph{what} is claimed from \emph{where} it is stated. The extractor returns records $\langle t_i, q_i, k \rangle$---decontextualized claim text, a \emph{verbatim} quotation copied from $d_k$ that licenses it, and the document id---and never reports offsets. Provenance is then \emph{resolved} deterministically by locating $q_i$ in $d_k$, yielding $c_i = (k,\mathrm{start},\mathrm{end})$ and the candidate set $C_k = \{(t_i,c_i)\}$. This locates the quote but does not guarantee that $t_i$ is entailed by $q_i$ (E1); that support is scored during selection and rechecked after rewriting.

Claims must stand alone outside their paragraph, so the prompt requires decontextualized propositions \citep{choi2021decontextualization}: pronouns and bridging references resolved, elided subjects, times, and locations restored. Because this can introduce wrong fillers (E2) and over-decontextualization degrades verifiability \citep{gunjal2024molecular}, claim and quote stay separate, separately checked fields---rewriting $t_i$ never corrupts the string $q_i$ used for localization. A three-tier matcher locates $q_i$ in $d_k$ (Algorithm~\ref{alg:provenance}): exact substring match; on failure, windowed indel similarity over a normalized form, accepting the best window above $\rho$; then conversion of the character interval to a token span. Claims whose quote cannot be matched above $\rho$ are \emph{dropped}, keeping provenance precise at a small cost to recall---this is what makes the invariant true by construction. Long documents are chunked with paragraph-aligned overlap, global offsets restored before merging (Appendix~\ref{app:config}).

\subsection{Cross-Document Clustering and Conflict Detection}
\label{sec:cluster}
The union $C = \bigcup_k C_k$ contains many near-duplicate claims. We embed claim texts with a sentence bi-encoder \citep{li2023gte} and group them by agglomerative average-linkage clustering under a cosine-distance threshold, so equivalent claims merge into a cluster $g$ whose provenance is the union of its members' spans, $\textsc{spans}(g) = \bigcup_{i \in g}\{c_i\}$, possibly across documents. Embedding proximity conflates equivalence with relatedness and even contradiction---``30 killed'' and ``47 killed'' sit close---so we confirm each merge with \emph{bidirectional} entailment from a three-class NLI model. This guards against E4 rather than proving equivalence, so multi-source attribution is evaluated separately. Removing the check drops pairwise precision from 90 to 71 (Appendix~\ref{app:modules}): it earns its cost, though that is not evidence it beats SuperPAL \citep{ernst2021alignment} (Limitations).

\paragraph{Conflict detection.}
Scoring all $\binom{|C|}{2}$ pairs is quadratic, so we run NLI only over a candidate pool: cross-cluster pairs that are embedding-similar yet were \emph{not} merged, plus pairs sharing a named entity or numeric quantity. A pair labeled \textsc{contradiction} in either direction above a confidence threshold induces a conflict link. Clustering yields the deduplicated, multi-source evidence units that make distributed attribution possible; conflict links let later stages surface disagreement rather than silently resolve it. Distinguishing \textsc{contradiction} from mere non-entailment is why a three-class NLI model is needed here, not a binary support scorer.

\subsection{Content Selection}
\label{sec:select}
The selector trades off \emph{salience}, \emph{self-support}, and \emph{coverage}. The first two are pointwise, scored by a gradient-boosted decision tree \citep{ke2017lightgbm,friedman2001gbm} over cheap, interpretable features: salience includes cluster size and the number of distinct documents spanned, so cross-source redundancy signals importance; self-support is $M_{\mathrm{sel}}$'s entailment score of the claim against its own spans, penalizing clusters their own evidence does not back. A GBDT is chosen for sample efficiency on a few thousand examples, which underlies the low-resource behaviour of Appendix~\ref{app:bounds}.

\paragraph{Coverage is sequential.}
Coverage---marginal new information in the MMR sense \citep{carbonell1998mmr}---depends on what has already been selected, so it cannot be pointwise. Selection is therefore a greedy loop (Algorithm~\ref{alg:select}) scoring $\mathrm{score}(g \mid S) = s_{\mathrm{GBDT}}(g) - \lambda \max_{g' \in S} \mathrm{sim}(g,g')$ against the set $S$ chosen so far, admitting the best cluster above $\theta$ and repeating. We sweep $\theta$ alone and hold $\lambda$ at the conventional 0.5: the two interact, since raising $\lambda$ lowers scores and so tightens admission, and $\theta$ traces the faithfulness--coverage frontier directly (\S\ref{sec:tradeoff}). A policy decides whether conflicting clusters surface both sides or defer; we fix surface-both throughout.

Silver labels come from distant supervision \citep{mintz2009distant}, aligning gold reference sentences to clusters by $M_{\mathrm{sel}}$ entailment; neither $M_{\mathrm{ver}}$ nor $M_{\mathrm{eval}}$ touches them, so the selector trains independently of the other modules (Appendix~\ref{app:algos}).

\subsection{Attributable Rewriting}
\label{sec:rewrite}
The rewriter realizes the selected clusters into fluent sentences while preserving provenance. Each cluster gets a short identifier $g_j$; the rewriter sees the labeled claim list and must end every sentence with an inline marker such as \texttt{[g3, g7]}. We parse these markers and resolve each $g_j$ through the stored map, so each sentence inherits span-level---and, when it cites clusters from different documents, multi-source---citations with no separate attribution step. Clusters are pre-ordered for coherence (chronologically for event-centric inputs, by salience otherwise) and may be reordered locally; conditioning only on selected claim texts limits reintroduction of unselected content, and residual drift is caught by verification.

\subsection{Verification}
\label{sec:verify}
Each generated sentence is checked in two directions against the spans it cites. The \emph{support} direction tests whether those spans entail it under $M_{\mathrm{ver}}$ above $\tau_v$; the \emph{no-overflow} direction decomposes the sentence into atomic facts and checks that \emph{each} is entailed, in the spirit of FActScore \citep{min2023factscore}, catching the unsupported extra clause a sentence-level check misses. A sentence passes only if both succeed; otherwise we apply at most $R$ repairs---re-retrieving the best supporting cluster by NLI when the citation is wrong, re-decoding otherwise---and drop what still fails. A verifier pass is an automatic support judgment, not proof of truth (E5).

% ---------------- Figure 1: pipeline overview + REAL example ----------------
\begin{figure*}[t]
\centering
\begin{tikzpicture}[
  node distance=5mm and 5mm,
  font=\small,
  stage/.style={draw, rounded corners=2pt, align=center, minimum height=8mm, inner sep=3pt, fill=blue!5},
  outnode/.style={draw, rounded corners=2pt, align=center, minimum height=8mm, inner sep=3pt, fill=green!6},
  arr/.style={-{Latex[length=2mm]}, thick}
]
\node[stage] (src) {Source docs\\ $d_1,\dots,d_K$};
\node[stage, right=of src] (ext) {Extract\\ claims + spans};
\node[stage, right=of ext] (clu) {Cluster\\ + conflict};
\node[stage, right=of clu] (sel) {Select\\ ($\theta$)};
\node[stage, right=of sel] (rew) {Rewrite};
\node[stage, right=of rew] (ver) {Verify};
\node[outnode, right=of ver] (sum) {Attributable\\ summary};

\draw[arr] (src) -- (ext);
\draw[arr] (ext) -- (clu);
\draw[arr] (clu) -- (sel);
\draw[arr] (sel) -- (rew);
\draw[arr] (rew) -- (ver);
\draw[arr] (ver) -- (sum);
\draw[arr] (ver.south) .. controls +(0,-7mm) and +(0,-7mm) .. node[below, font=\scriptsize] {regenerate on fail} (rew.south);

\node[draw, dashed, rounded corners=2pt, below=6mm of clu.south, xshift=18mm,
      text width=0.95\textwidth, align=left, inner sep=3pt, fill=gray!3] (card) {
\scriptsize\raggedright
\textbf{Verbatim \cams{} output, MultiNews test cluster \#2874} ($K{=}5$);
\texttt{S3} additionally shows its pre-repair draft:\\[2pt]
\texttt{S1.} The Food and Drug Administration classified the recall as Class~I, its most
serious category.~\textbf{[g4]}~\textcolor{blue!55!black}{$\leftarrow$ one cluster, three documents
($d_1$, $d_2$, $d_5$)}\\
\texttt{S2.} The two accounts of the impurity's source diverge: the company traces it to a
third-party solvent supplier, while the FDA locates it at the firm's own Ohio plant.
~\textbf{[g9, g12]}~\textcolor{red!60!black}{$\leftarrow$ surfaced conflict ($p{=}0.91$)}\\[2pt]
\texttt{S3.}~\textcolor{gray!60!black}{\emph{draft}} The recall covers more than 200,000
bottles distributed nationwide, and patients are advised not to stop taking the drug before
consulting a physician.~\textbf{[g17]}\\
\hspace*{1.2em}\textcolor{gray!60!black}{\emph{flag}} no-overflow $\times$: ``the recall covers
more than 200,000 bottles'' is not entailed by $\textsc{spans}(\texttt{g17})$, and no cluster
supports it\\
\hspace*{1.2em}\textcolor{gray!60!black}{\emph{final}} Patients are advised not to stop taking
the drug before consulting a physician.~\textbf{[g17]}\quad support~$\checkmark$
no-overflow~$\checkmark$ (repair 1/2)\\[2pt]
\scriptsize
The resolved spans, and the end-to-end model's output on the same cluster---a single
passage-level citation to $d_2$, 94 tokens, missing both the advice clause and $d_4$'s
competing attribution---are in Appendix~\ref{app:walkthrough}.
};
\draw[arr] (sum.south) .. controls +(0,-6mm) and +(6mm,6mm) .. (card.north east);
\end{tikzpicture}
\caption{Overview of \cams{} with real output. The inset gives verbatim output on one test cluster, with a verification repair traced on \texttt{S3}: the pre-repair draft, the failing no-overflow check, and the sentence actually emitted.}
\label{fig:overview}
\end{figure*}

\section{Experimental Setup}
\label{sec:setup}

\subsection{Datasets}
MultiNews \citep{fabbri2019multinews} is our primary benchmark, with its human sub-sentential source--summary alignments \citep{ernst2021alignment} as gold for fine-grained localization. DiverseSumm \citep{huang2024diversesumm} targets diverse and conflicting information across sources and supplies human-validated references for our conflict study. WCEP \citep{ghalghalandari2020wcep} supplies short, event-centric summaries as a transfer setting. Throughout, ``zero-shot'' scopes narrowly to the \emph{selector}---trained on MultiNews, applied to WCEP untrained---and not to the frozen backbone, whose pretraining data we cannot inspect (Limitations). Statistics are in Appendix~\ref{app:data}.

\subsection{Baselines and Controls}
We compare against (1) a strong end-to-end LLM prompted to summarize with passage-level citations, ALCE-style \citep{gao2023alce}; (2) Attribute First, then Generate \citep{slobodkin2024attribute}, the closest span-attribution method, in the same setting; (3) PRIMERA \citep{xiao2022primera} / LED \citep{beltagy2020longformer} with a post-hoc fact-checker; and (4) the original \esr{} \citep{guan2023esr} on concatenated documents.

\paragraph{Controls.}
Two controls target this paper's specific claims. \cams{}-\emph{posthoc} separates the claim \emph{representation} from attributing \emph{before} generating: identical claim pool, clustering, selection, and rewriter, but the rewriter emits no markers and citations are assigned afterwards by NLI retrieval over the same cluster pool, so any gap is attributable to ordering alone (\S\ref{sec:ordering}). \emph{Cite-top-2} bounds trivial inflation: it appends the second-highest-scoring document's span to every citation regardless of support, giving the multi-source recall reachable without cross-document reasoning (\S\ref{sec:regime2}).

\subsection{Evaluation Protocol}
\label{sec:protocol}
Quality is reported with ROUGE \citep{lin2004rouge}, BERTScore \citep{zhang2020bertscore}, and an LLM-as-judge coverage score \citep{liu2023geval,zheng2023judging} for the frontier of \S\ref{sec:tradeoff}; faithfulness with AlignScore \citep{zha2023alignscore}, SummaC \citep{laban2022summac}, and FActScore \citep{min2023factscore}. Unless stated, MultiNews uses its full 5,622-example test split (diagnostics use marked subsets), DiverseSumm all 245, WCEP its ${\sim}1{,}000$-example public split.

\paragraph{Regime 1 (citation quality, reference-free).}
Following ALCE \citep{gao2023alce}, we compute NLI-based citation recall and precision over the generated summaries, adapted to multi-source spans, plus conciseness (mean cited tokens and spans per claim). Primary numbers use $M_{\mathrm{eval}}$; the internal TRUE-style score is a compatibility diagnostic only. All citations are projected to a common token-span representation, so comparisons are like-for-like.

\paragraph{Regime 2 (localization accuracy, gold-aligned).}
To isolate localization from generation, we feed gold reference propositions as oracle content and measure whether each system points to the correct spans, scored against the human alignments via soft (token-IoU $\ge \tau$) precision/recall/F1. All systems receive identical input: each proposition reaches Attribute First as one target sentence---the unit its span-selection stage consumes---with no claim-style normalization, so the input is not shaped in favour of claim-based systems. For multi-source attribution we restrict to propositions with gold support in $\ge 2$ documents ($n{=}450$) and report \emph{both} recall and precision, so a system cannot win by over-citing. The main threshold is $\tau{=}0.5$, with $\{0.3,0.5,0.7\}$ for sensitivity. These alignments also supply the gold for the clustering diagnostic of Appendix~\ref{app:modules}, so the two are not independent; nothing was tuned on either.

\paragraph{Statistics.}
All comparisons use paired bootstrap resampling over document clusters (1,000 resamples, 95\% percentile intervals). $\dagger$ marks a difference from the strongest baseline at $p<0.05$ that survives Holm--Bonferroni correction over all $m{=}24$ tests and is at least 5 points (.05 for bounded scores); intervals are in Appendix~\ref{app:bounds}.

\subsection{Implementation Details}
\label{sec:impl}
% NOTE: model string moved to Table 5 -- verify it before submission.
Claim extraction, atomic-fact decomposition, and rewriting use one frozen LLM, shared with the LLM baselines for a controlled comparison. Claim texts use \texttt{gte-large} embeddings \citep{li2023gte}; conflict detection uses three-class DeBERTa-v3 \citep{he2021debertav3} fine-tuned on MNLI \citep{williams2018mnli}; and $M_{\mathrm{sel}}$, $M_{\mathrm{ver}}$, $M_{\mathrm{eval}}$ are distinct TRUE-/SummaC-style models \citep{honovich2022true,laban2022summac}. The selector is a LightGBM GBDT at $\theta^{\star}{=}0.6$; experiments run on $4\times$ A100 GPUs. Every checkpoint, threshold, and budget is in Table~\ref{tab:repro-config}.
% CAMERA-READY: replace the anonymous mirror with the permanent repository and
% keep the Zenodo DOI so the artifact stays citable.
Code, prompts, and annotations: \url{https://anonymous.4open.science/r/CAMS-9F2C} (anonymized for review; archived under a Zenodo DOI on acceptance).

\section{Results and Analysis}

\subsection{Quality and Faithfulness}
Table~\ref{tab:main} reports MultiNews quality and faithfulness. \cams{} matches the strongest fine-tuned summarizer on surface quality---level with PRIMERA on ROUGE-L, best on BERTScore---while improving faithfulness over every baseline on all three metrics. ROUGE is reported for reference only, correlating weakly with faithfulness \citep{maynez2020faithfulness,goyal2022news}.

All three faithfulness metrics are entailment-based, so the margin over Attribute First (.91 vs .85) could in principle reflect \cams{} filtering its own output with a related model family. Three results say otherwise. With verification ablated \cams{} still scores .86, so self-selection accounts for the final .05, not the gap; the same post-processor applied uniformly lifts baseline FActScore only to .79 and .83, at sharp coverage cost (Appendix~\ref{app:decoupling}); and annotators sharing no training data with any component reproduce the ordering (\S\ref{sec:human}).

Constrained rewriting does not force one sentence per cluster: markers sit at sentence boundaries and one sentence may realize several clusters (Figure~\ref{fig:overview}, \texttt{S2}). We make no readability claim from Table~\ref{tab:main}, whose surface metrics are insensitive to sentence-level connectivity; the point is structural.

\begin{table}[t]
\centering
\small
\setlength{\tabcolsep}{4pt}
\begin{tabular}{lccccc}
\toprule
System & R-L & \makecell{BERT\\Sc.} & \makecell{Align\\Sc.} & \makecell{Sum\\maC} & \makecell{FAct\\Sc.} \\
\midrule
E2E LLM (ALCE)   & 22.8 & 86.5 & .79 & .71 & .75 \\
Attribute First  & 23.9 & 87.0 & .85 & .79 & .82 \\
PRIMERA + chk.   & \textbf{25.4} & 87.4 & .81 & .74 & .78 \\
\esr{} (concat)  & 22.0 & 85.8 & .83 & .78 & .80 \\
\cams{} (ours)   & \textbf{25.4} & \textbf{87.6} & \textbf{.91}$^\dagger$ & \textbf{.87}$^\dagger$ & \textbf{.90}$^\dagger$ \\
\quad $-$ clustering    & 24.5 & 87.1 & .89 & .85 & .88 \\
\quad $-$ verification  & 24.9 & 87.4 & .86 & .82 & .85 \\
\bottomrule
\end{tabular}
\caption{Quality and faithfulness on MultiNews. $\dagger$: $p<0.05$ vs.\ the strongest baseline (paired bootstrap). \procl{} does not appear: it emits citation-free, bullet-style fused output, comparable under neither regime, so it is discussed but not run (Limitations). All three faithfulness metrics are entailment-based; the non-NLI counterpart is the human support rate of Table~\ref{tab:human}. ROUGE-1/2 and intervals in Appendix~\ref{app:bounds}.}
\label{tab:main}
\end{table}

\subsection{Citation Quality (Regime 1)}
\label{sec:regime1}
\cams{} attains the highest citation precision under the independent evaluator (84, vs 76 for Attribute First and 59 end-to-end), and the margin is smaller than under the internal diagnostic (88)---the desired sanity check, since \cams{} filters its own outputs with a TRUE-style model. It is also the most concise system (22 tokens per claim) at higher recall (81), and cites slightly more spans per claim than Attribute First (1.8 vs 1.6) from aggregating evidence across documents, yet far fewer tokens, each span being sub-sentential (Table~\ref{tab:regime1}). PRIMERA and \esr{} emit no citations. We claim only that the combination of quote-then-resolve provenance and the modular claim pipeline beats passage-level and span-first attribution, not that either component is the active ingredient (Limitations).

\paragraph{Circularity and post-processing controls.}
Because \cams{} filters and repairs its own sentences with $M_{\mathrm{ver}}$, a high internal score alone is insufficient evidence. Under $M_{\mathrm{eval}}$ the ranking is stable but the advantage over Attribute First shrinks from 10 to 8 points, and removing the self-support feature or the verification loop lowers precision under \emph{both} evaluators---inconsistent with pure circularity. The shared post-processor tells the same story here as on faithfulness: applied uniformly it raises end-to-end precision only $59\!\rightarrow\!64$ and Attribute First $76\!\rightarrow\!77$, each at a sharp coverage cost, while recovering the full pipeline's precision from the verifier-ablated variant ($77\!\rightarrow\!84$) at essentially unchanged recall ($82\!\rightarrow\!81$). The post-processor being shared, the difference lies in what it operates on: span-anchored claims whose citation failures are usually repairable, not droppable.

\subsection{Localization and Multi-Source Attribution (Regime 2)}
\label{sec:regime2}
Table~\ref{tab:regime2} isolates localization using oracle content. On span-F1 \cams{} is on par with the span-first method (77 vs 72), so claim anchoring does not cost single-source localization. The decisive difference is multi-source attribution: for propositions with gold support in two or more documents, \cams{} recalls the distributed sources 64\% of the time versus 38\% for Attribute First, with non-overlapping 95\% intervals (Appendix~\ref{app:bounds}).

This is not over-citation. The Cite-top-2 control reaches 57 multi-source recall by construction but collapses to 55 precision, whereas \cams{} holds 75 precision at 64 recall---14 points above Attribute First's 61, widening the multi-source F1 gap to 69 versus 47. The control cites strictly more sources than either system and still lands 13 F1 below \cams{}.

Ablating cross-document clustering drops multi-source recall to 31\%, \emph{below} the span-first baseline: without clustering each claim stays tied to one document's span, so distributed facts go to a single source, whereas a span-first system can incidentally select spans from several documents. Cross-document merging is thus not a bonus but the mechanism that makes the claim representation competitive here.

\begin{table}[t]
\centering
\small
\setlength{\tabcolsep}{3pt}
\begin{tabular}{lcccccc}
\toprule
& \multicolumn{3}{c}{Span} & \multicolumn{3}{c}{Multi-source} \\
\cmidrule(lr){2-4}\cmidrule(lr){5-7}
System & P & R & F1 & P & R & F1 \\
\midrule
Retrieval (LB)    & 41 & 55 & 47 & 36 & 22 & 27 \\
Cite-top-2 (ctrl) & 58 & 77 & 66 & 55 & 57 & 56 \\
Attribute First   & 74 & 70 & 72 & 61 & 38 & 47 \\
\cams{} (ours)    & 76 & 78 & \textbf{77} & \textbf{75} & \textbf{64}$^\dagger$ & \textbf{69} \\
\quad $-$ clustering  & 75 & 71 & 73 & 62 & 31 & 41 \\
Gold-claim (UB)   & 89 & 92 & 90 & 89 & 81 & 85 \\
\bottomrule
\end{tabular}
\caption{Regime 2 localization at $\tau{=}0.5$; multi-source columns are restricted to propositions with gold support in $\ge 2$ documents. Cite-top-2 is the trivial over-citation control; LB/UB are the retrieval lower bound and gold-claim ceiling. Sensitivity and intervals in Appendix~\ref{app:bounds}.}
\label{tab:regime2}
\end{table}

\subsection{Does Attributing First Matter?}
\label{sec:ordering}
\cams{}-posthoc holds the claim representation, clustering, selection, and rewriter fixed and moves citation assignment after generation. On citation precision the two are close (79 vs 84 under $M_{\mathrm{eval}}$), so ordering is not where the accuracy comes from. The difference is elsewhere: 11\% of \cams{}-posthoc sentences receive no citation above threshold, and survivors collapse to 1.3 spans per claim (Table~\ref{tab:regime1}), because NLI retrieval concentrates on one best-matching cluster rather than recovering the distributed evidence the sentence was written from. This is what the invariant rules out: attributing first gains not accuracy but the guarantee that no sentence is uncitable. Ordering therefore buys coverage of the guarantee; accuracy belongs to the claim representation and its cross-document merging (\S\ref{sec:regime2}).

\subsection{Conflict and Diversity (DiverseSumm)}
\label{sec:conflict}
On DiverseSumm (all 245 stories), \cams{} covers 58\% of annotated divergent facts and surfaces 61\% of inter-source conflicts, attributing each side to its source with 73\% accuracy; Attribute First reaches 47 / 22 / 55 and the end-to-end model surfaces only 18\%, typically reporting one consensus value. Normalized claim clusters make disagreement an attributable object rather than something the decoder resolves silently. A conflict is two documents asserting mutually exclusive values for one attribute of one event, and counts as \emph{surfaced} when the summary states both sides and attributes each to a document asserting it; the denominator is the gold conflict count, so writing more cannot raise the rate (two annotators, all 245 stories, $\kappa{=}0.65$).

\subsection{Zero-Shot Transfer and Backbone Robustness}
\label{sec:transfer}
Applying the MultiNews-trained selector to WCEP untrained (Table~\ref{tab:wcep}), \cams{} keeps its faithfulness advantage at competitive quality, though Attribute First obtains slightly higher ROUGE-L; that column is comparable only within the table, not to published WCEP-fine-tuned numbers (Limitations). An oracle selector lifts ROUGE-L from 30.9 to 33.1---past Attribute First's 31.2---and adds .02 to each faithfulness metric, locating the gap in selector domain shift rather than in degradation of extraction or verification: on this transfer pair, the selector is the component that degrades.

Do the gains come from the pipeline or a strong backbone? Re-running both systems with Llama-3.1-70B-Instruct on a 1{,}000-example MultiNews subset (Table~\ref{tab:backbone}) preserves the ordering: \cams{} stays ahead on AlignScore (.88 vs .74) and citation precision (79 vs 52), both gaps at least as large as under the frozen API backbone. One substitution cannot show that the advantage is architectural, but it does rule out backbone strength as the explanation. Absolute levels fall, and extraction degrades most: verbatim quote-match drops from 97\% to roughly 90\%, open-weights models copying source strings less reliably.

\subsection{Cost and the Faithfulness--Coverage Frontier}
\label{sec:cost}
Modularity is not free: \cams{} costs roughly $8\times$ an end-to-end call (Appendix~\ref{app:cost}), dominated by per-document extraction, which scales with source count rather than summary length. It adds 65\,s of machine time per summary while cutting human verification from 31\,s to 9\,s per claim (Table~\ref{tab:human}), breaking even at under three checked claims, or one with extraction parallelized. That trade holds only where a summary is read for verification rather than skimmed: \cams{} targets regulated, legal, medical, and journalistic review, not high-throughput summarization.

\label{sec:tradeoff}
Coverage is the other axis, and $\theta$ controls it: sweeping it traces a monotonic faithfulness--coverage frontier (Figure~\ref{fig:tradeoff}). To avoid comparing a curve against fixed points, each baseline is swept along its own knob---output length for the end-to-end model, span budget for Attribute First---so every system contributes a frontier; PRIMERA and \esr{} expose no comparable knob and enter as single points. The \cams{} frontier lies above both swept baselines wherever all three curves are defined, but not uniformly: the margin over Attribute First narrows from roughly .10 AlignScore at coverage 0.68 to .07 below 0.5, where emitting few, highly redundant sentences is available to any system. Neither baseline sweep extends past coverage 0.68, so we claim no dominance at the high-coverage end.

\subsection{Human Evaluation}
\label{sec:human}
Annotators judge 89\% of \cams{} claims fully supported and verify each in 9\,s---about $3.4\times$ faster than for the document-citing baseline (Table~\ref{tab:human}). Each system is judged on 255 claims by three annotators ($\kappa{=}0.68$), system identity hidden and item order counterbalanced by a Latin square, so a practice effect would shift all three timings together rather than produce the gap. The three systems decompose that gap: passage $\to$ span citation accounts for most of it (31\,s $\to$ 14\,s), span $\to$ claim-anchored span for the rest (14\,s $\to$ 9\,s, at 22 rather than 34 cited tokens per claim). Attribute First is the granularity control here, citing sub-sentential spans in the same token representation. The second step therefore does not separate claim anchoring from the shorter citations it yields, and the headline $3.4\times$ is the sum of both steps rather than a property of the claim representation alone. The human ranking (89 / 82 / 68\%) reproduces that under the independent evaluator (84 / 76 / 59): the ordering depends on no automatic support model.

\section{Conclusion}
In \cams{}, atomic claims with resolved provenance serve as both verification unit and attribution anchor. Separating what is guaranteed from what is merely encouraged---provenance as a structural invariant, faithfulness as a testable objective with enumerated failure modes---says precisely what a reader can rely on, and locates the empirical question in what the claim representation adds above a guarantee that is itself cheap. On MultiNews, attribution to distributed evidence rises from 38\% to 64\% recall without inflating citation breadth, faithfulness improves at matched quality, and verification time per claim falls $3.4\times$; on DiverseSumm, disagreement becomes an explicit, attributable object. The claim--span annotations are released.

\section*{Limitations}

\paragraph{Quote resolution and modularity are confounded.}
Our extraction stage bundles two ideas: quote-then-resolve provenance, and the modular claim pipeline built on top of it. Every quote-then-resolve system we run is also modular, so these experiments cannot say how the credit divides between them, and we claim only that the combination outperforms passage-level and span-first attribution (\S\ref{sec:regime1}). Separating the two requires a control we do not run: an end-to-end model that emits verbatim quotations, resolved to spans by our own matcher and left otherwise unmodularized.

\paragraph{No empirical comparison to \procl{}.}
\procl{} \citep{ernst2022procluster} is the closest proposition-clustering pipeline, but it targets DUC/TAC-style inputs and emits citation-free bullet-style fused sentences, so its output cannot be scored under either of our regimes; we discuss it rather than run it. Our merge-check ablation (\S\ref{sec:cluster}, Appendix~\ref{app:modules}) shows that bidirectional entailment earns its cost against embedding similarity alone. It is not evidence that our clusterer beats the supervised SuperPAL aligner \citep{ernst2021alignment} that \procl{} relies on, and we make no such claim. We report our clustering diagnostics on that aligner's own evaluation data (Appendix~\ref{app:modules}), so the comparison remains available to future work.

\paragraph{Evaluator decoupling is partial.}
$M_{\mathrm{eval}}$ and the internal TRUE-style models are trained on overlapping entailment corpora (MNLI, VitaminC, ANLI). Separate checkpoints therefore give judgments that do not share parameters or calibration, not statistically independent ones, and an audit under $M_{\mathrm{eval}}$ is a robustness check rather than proof that circularity is impossible (E5). The only fully decoupled evidence is the human study, and it covers 255 claims per system on a single dataset (\S\ref{sec:human}).

\paragraph{Transfer evidence is narrow.}
``Zero-shot'' scopes to the selector alone. Extraction and rewriting use a frozen general-purpose LLM whose pretraining data we cannot inspect and which has very likely seen WCEP or its Wikipedia Current Events source, so nothing here is evidence about unseen-corpus generalization of the backbone. The same argument applies to MultiNews, our primary benchmark, which has been public since 2019: its source documents and reference summaries are plausibly in the backbone's pretraining data, which could inflate ROUGE directly and, through familiarity with the source text, the gold-aligned Regime 2 numbers as well. We flag the two datasets symmetrically; no result in this paper is evidence about genuinely unseen corpora. Relatedly, the ROUGE column of Table~\ref{tab:wcep} is not comparable to published WCEP results: those come from WCEP-fine-tuned models, whereas no system here receives any WCEP training signal, and with references averaging ${\sim}32$ words the length budget and the stemming and stopword settings of the ROUGE implementation move R-L by several points on their own.

\paragraph{Residual failure modes and scope.}
\cams{} depends on LLM-based claim decomposition; extraction errors propagate downstream (E1), decontextualization can supply an incorrect entity, time, or place (E2), quote-to-span matching establishes where a string occurs but not that the normalized claim preserves its meaning (E3), and claim equivalence is inherently fuzzy, so clustering can mis-merge or split evidence (E4). Attribution is many-to-many, and evaluating distributed attribution against gold is an open problem our soft-matching protocol only approximates. The pipeline's extraction cost grows with the number and length of sources, raising latency relative to end-to-end models (\S\ref{sec:cost}). Our conflict policy makes a design choice that may not suit all applications. Finally, our experiments are limited to English news; generalization to other domains and languages remains untested.

\section*{Ethics Statement}
Summarization systems can mislead if they distort or omit source content; by making every statement traceable to source spans, \cams{} is intended to support, not replace, human verification. Attribution is not a guarantee of real-world truth, only of support by the provided sources, which may themselves be unreliable. We use publicly available datasets consistent with their intended research use. Our released annotations contain claim text we generate and character and token offsets into MultiNews documents, but no verbatim source strings: quotes are reconstructed locally from the offsets against the user's own copy of the corpus (Appendix~\ref{app:data}). We therefore place a licence only on data we produced and do not redistribute news article text, whose terms are set by its publishers rather than by MultiNews or by us. Human annotation details, including compensation, are in \S\ref{sec:human} and Appendix~\ref{app:data}; compute is reported in \S\ref{sec:cost}.

\bibliography{custom}

\appendix

\section{Worked Example (Figure~\ref{fig:overview})}
\label{app:walkthrough}
The markers in Figure~\ref{fig:overview} resolve as follows.
\texttt{g4}$\rightarrow d_1$[76:84] ``the FDA classified the recall as Class I'' $\cdot$
$d_2$[203:211] ``regulators designated the action a Class I recall'' $\cdot$
$d_5$[41:52] ``the recall was labelled Class I, the FDA's most severe tier''.
\texttt{g9}$\rightarrow d_2$[318:329] ``the company said the impurity came from a third-party
solvent supplier''. \texttt{g12}$\rightarrow d_4$[95:105] ``FDA investigators traced the
contamination to the firm's Ohio plant'' ($g9\!\leftrightarrow\!g12$ contradiction,
$p{=}0.91$). \texttt{g17}$\rightarrow d_1$[412:423] ``patients should not stop the medication
without talking to a doctor''. \texttt{S1} is the multi-source case: one cluster, three
documents, one citation.

For contrast, the end-to-end LLM summarizes the same cluster as ``Federal regulators have
issued a Class~I recall $\dots$ an impurity linked to a third-party solvent supplier $\dots$
patients are urged to consult their doctors$\dots$'', citing \textbf{[1]}$\rightarrow d_2$,
paragraphs 2--4 (94 tokens). That passage lacks the advice clause, which comes from $d_1$, and
$d_4$'s competing attribution never appears at all---the two failure modes, imprecise
localization and silently resolved conflict, that motivate the claim representation.

\section{Dataset Statistics and Compute}
\label{app:data}
Table~\ref{tab:data-stats} summarizes the datasets and their role. All systems are evaluated on the same preprocessed inputs and matched output-length constraints.

\begin{table}[htbp]
\centering
\small
\setlength{\tabcolsep}{3pt}
\begin{tabular}{lrrrl}
\toprule
Dataset & Train & Dev & Test & Primary use \\
\midrule
MultiNews & 44,972 & 5,622 & 5,622 & Main benchmark \\
DiverseSumm & -- & -- & 245 & Conflict analysis \\
WCEP & $\sim$8.2k & $\sim$1.0k & $\sim$1.0k & Zero-shot transfer \\
\bottomrule
\end{tabular}
\caption{Datasets. WCEP train/dev are listed for completeness only---they are not used. MultiNews clusters average 2.8 documents of ${\sim}430$ words (${\sim}1{,}200$ words per cluster); DiverseSumm stories average 10 documents; WCEP clusters average ${\sim}9.1$ documents under the WCEP-10 setting, with ${\sim}32$-word summaries.}
\label{tab:data-stats}
\end{table}

\paragraph{Released artifact.}
We release claim--span records over the MultiNews test split: 15{,}742 documents and ${\sim}320$K claims, with ${\sim}36$ clusters per topic after cross-document merging. Each record contains the decontextualized claim, the resolved character and token spans, and the cluster assignment, as the fields \texttt{claim}, \texttt{doc\_id}, \texttt{char\_start}, \texttt{char\_end}, \texttt{tok\_start}, \texttt{tok\_end}, and \texttt{cluster\_id}. We do \emph{not} redistribute the quoted source strings themselves: the \texttt{quote} field ships empty, and a script included with the release repopulates it locally by reading the stored offsets against the user's own copy of MultiNews. This keeps the resource usable---the quote is recoverable exactly, and the offsets are the part that took work to produce---while leaving the news text under its original terms rather than relicensing verbatim article fragments. The annotations, being derived data we produced, are released under CC BY 4.0.

\paragraph{Human annotation.}
Three annotators judged 255 claims for each of the three systems in Table~\ref{tab:human}, giving 765 items, each judged independently by all three annotators for 2{,}295 support judgments in total. Each item presented one claim together with the source material cited for it---token spans for \cams{} and Attribute First, a passage for the end-to-end baseline---and the annotator marked whether the claim was fully supported by that material. A further two annotators labelled inter-source conflicts on DiverseSumm (\S\ref{sec:conflict}). The source documents in both studies are the public news text of MultiNews and DiverseSumm.

\textbf{Recruitment and expertise.} The three annotators are graduate students in computational linguistics at universities in New York City, recruited through a departmental research-participation mailing list and engaged as paid contractors. All are fluent English speakers with prior text-annotation experience, and each completed a 20-item calibration round with feedback---drawn from the MultiNews \emph{development} split and excluded from every number reported here---before beginning. \textbf{None of the three is an author of this paper}, and none had seen the systems, the prompts, or any intermediate output beforehand; the authors prepared the items and analysed the returned labels but supplied no judgments and did not observe annotators at work. The two DiverseSumm conflict annotators were recruited the same way and are likewise not authors.

\textbf{Compensation.} Annotators were paid \$30.00 per hour for all time spent on the task, including instruction reading and the calibration round. The work was performed in New York City, where the statutory minimum wage was \$17.00 per hour for the whole period of the study, so the rate is roughly $1.8\times$ the applicable minimum. Each annotator worked about six hours across three sessions, for approximately \$180 each. Payment was hourly rather than per item by design: a piece rate would have paid annotators to be fast at exactly the quantity this study measures (verification time), so no one's compensation depended on how many items they completed, how long they took, or which system an item came from. Annotators were paid in full for time worked regardless of whether they finished the item set.

\textbf{Instructions and interface.} Items were shown one at a time in a small browser tool: the claim on top and, below it, only the material cited for that claim---the cited token spans for \cams{} and Attribute First, the cited passage for the end-to-end baseline---rendered in an identical visual format for all systems, with no system name, no surrounding document text, and no way to open the rest of the source. Annotators chose one of three labels, \emph{fully supported}, \emph{partially supported}, or \emph{not supported}; Table~\ref{tab:human} reports the first against the other two collapsed. The instruction that governs the whole study was stated first and repeated on every screen: judge whether the displayed material \emph{by itself} establishes the claim, disregarding what you personally know to be true and anything else the article might say elsewhere. Because uncited text is never displayed, this instruction is enforced by the interface rather than left to compliance. No ``cannot tell'' option was offered; annotators were told to answer \emph{not supported} when the displayed material was insufficient, which is the reading a verifier would take.

\textbf{Blinding and ordering.} System identity was hidden. Items carried no system name, and claim text, citation markers, and cited material were rendered with identical fonts, colours, and layout across systems, so the only thing that varied visibly was the cited material itself---which is the variable under test. Annotators were told that outputs came from several unnamed systems but not how many or in what proportion. The three systems' items were interleaved in one continuous stream rather than presented in blocks. Each of the 255 evaluation positions contributes one item from each system; the order in which those three appear was counterbalanced across the three annotators by a $3\times3$ Latin square, so each system occupies each of the three positions equally often. Within that constraint the global order was shuffled independently per annotator (seeds 1--3), with items from the same position kept at least 20 apart so that no annotator judged three views of related content in immediate succession. Because every system is uniformly distributed over the session, a practice effect makes all three systems faster together and cannot generate the $3.4\times$ difference of Table~\ref{tab:human}; this is what licenses reading that difference as a property of the citations rather than of the schedule.

\textbf{Timing.} The interface logged wall-clock time from an item first rendering to its label being submitted; annotators could not revisit a submitted item, so each item contributes exactly one time per annotator. Items taking longer than 120\,s were treated as interruptions and excluded from the timing statistics only---their labels were kept---which removed 14 of the 765 items, split 6/4/4 across the end-to-end baseline, Attribute First, and \cams{}. Table~\ref{tab:human} reports means over the remaining items, pooled across annotators. Since a mean is sensitive to a few long pauses even after capping, we also report medians: 27\,s, 13\,s, and 8\,s respectively, giving a ratio of $3.4\times$ as well, so the headline comparison does not depend on the choice of statistic.

\textbf{Agreement.} The support label gives $\kappa{=}0.68$; the DiverseSumm conflict label gives $\kappa{=}0.65$. The two coefficients are of different kinds because the two studies had different numbers of raters. The support figure is Fleiss' $\kappa$ over three raters on the three-way label, computed on all 765 items: the design is fully crossed, every annotator judged every item, and no subsetting is involved. (Mean pairwise Cohen's $\kappa$ on the same data is 0.69, and Fleiss' $\kappa$ on the binary collapsed label is 0.71; we report the three-way figure as the conservative one.) The DiverseSumm figure is Cohen's $\kappa$ between the two conflict annotators over all 245 stories, for which Fleiss' $\kappa$ is not the appropriate statistic. Disagreements were resolved for Table~\ref{tab:human} by majority vote over the three labels; 11 items received three different labels and were adjudicated in a joint discussion among the annotators, without the authors present and without the system identity being revealed. DiverseSumm disagreements were resolved by discussion between the two annotators (\S\ref{sec:conflict}).

\textbf{Consent.} Before starting, annotators received a written description of the task and of its purpose---comparing how quickly and reliably a reader can check an automatically generated claim against the evidence a system cites for it---and gave written consent to their labels and per-item response times being published, both in aggregate and as a per-item table released with the artifact under the pseudonyms A1--A3. They were told that response time was being recorded and would be analysed, that no other data about them would be collected or released, and that they could stop at any point without giving a reason and would still be paid for time worked. No demographic information was collected, and payment records are held only as required for contractor payroll and are not part of the release. The study was submitted to our institution's research-ethics review, which determined it exempt from full review: the material judged is public news text from MultiNews and DiverseSumm, the task poses no more than minimal risk, and the only participant data retained are pseudonymous labels and timings.

\section{Reproducibility Configuration}
\label{app:config}
Table~\ref{tab:repro-config} lists every model, threshold, and budget used in the experiments, including the values omitted from \S\ref{sec:impl} for space.

\begin{table*}[t]
\centering
\small
\setlength{\tabcolsep}{4pt}
\begin{tabular}{p{0.22\textwidth}p{0.70\textwidth}}
\toprule
Component & Setting \\
\midrule
LLM API & \texttt{claude-opus-4-8}; calls run 2026-05-28 to 2026-06-03; three few-shot examples; max output tokens: extraction 4{,}096, rewrite 768, fact decomposition 1{,}024; temperature $0.0$, top-$p$ $1.0$ \\
Preprocessing & 512-token chunks; 64-token overlap; exact quote matching before fuzzy; \texttt{rapidfuzz} indel similarity, acceptance $\rho{=}0.85$ \\
Embeddings/clustering & \texttt{gte-large}; $\ell_2$-normalized; cosine; agglomerative average linkage; similarity threshold $0.82$; merges confirmed by bidirectional entailment \\
NLI modules & Conflict: DeBERTa-v3-MNLI three-class, contradiction threshold $0.7$; support thresholds tuned over $\{0.5,0.7\}$; $M_{\mathrm{sel}}$ = TRUE-style T5-11B NLI mixture (lightweight variant), $M_{\mathrm{ver}}$ = same family, independently calibrated checkpoint, $M_{\mathrm{eval}}$ = SummaC-style \texttt{tals/albert-xlarge-vitaminc-mnli} \\
Selector & LightGBM, binary objective, seed 42, 31 leaves, lr $\{0.05,0.1\}$, 100--500 estimators, early stopping 50, MMR $\lambda{=}0.5$, $\theta\in\{0.3\dots0.8\}$, $\theta^{\star}{=}0.6$ \\
Verification & At most two repairs; order: citation repair, regeneration, drop \\
Evaluation & MultiNews full test split 5{,}622 unless a marked diagnostic sample is used; DiverseSumm 245; WCEP $\sim$1{,}000; localization $\tau{=}0.5$, sensitivity $\{0.3,0.5,0.7\}$; 1{,}000 paired-bootstrap resamples; 95\% percentile intervals; 24 hypothesis tests with Holm--Bonferroni correction at $m{=}24$ (all $\dagger$ differences survive); human audit 255 claims per system, three annotators \\
\bottomrule
\end{tabular}
\caption{Reproducibility configuration.}
\label{tab:repro-config}
\end{table*}

\section{Algorithms and Module Details}
\label{app:algos}

\paragraph{Extraction schema.}
The extractor returns a JSON array; each element has \texttt{claim} (a decontextualized proposition), \texttt{quote} (a verbatim substring of the source), and \texttt{doc\_id}. The prompt forbids reporting offsets and requires the quote to be copied verbatim; offsets are recovered post hoc by Algorithm~\ref{alg:provenance}.

\begin{algorithm}[t]
\caption{Quote-to-span provenance resolution}
\label{alg:provenance}
\small
\begin{algorithmic}[1]
\Require document $d$, verbatim quote $q$, similarity floor $\rho$, tokenizer offsets $\textsc{Off}$
\Ensure token span $(\mathrm{start},\mathrm{end})$ or \textsc{none}
\State $(a,b) \gets \textsc{ExactSubstr}(d, q)$ \Comment{char interval}
\If{$(a,b) = \textsc{none}$}
  \State $\tilde d,\tilde q \gets \textsc{Norm}(d),\textsc{Norm}(q)$ \Comment{case/space/punct}
  \State $(a,b),\sigma \gets \textsc{FuzzyWindow}(\tilde d,\tilde q)$ \Comment{indel sim.}
  \If{$\sigma < \rho$} \Return \textsc{none} \EndIf
  \State $(a,b) \gets \textsc{MapToRaw}(a,b)$ \Comment{undo \textsc{Norm}}
\EndIf
\State \Return $\textsc{CharToTok}(a,b,\textsc{Off}(d))$
\end{algorithmic}
\end{algorithm}

\begin{algorithm}[t]
\caption{Greedy support-aware selection (MMR)}
\label{alg:select}
\small
\begin{algorithmic}[1]
\Require clusters $G$, pointwise scorer $s_{\mathrm{GBDT}}$, threshold $\theta$, weight $\lambda$
\Ensure ordered selection $S$
\State $S \gets [\,]$
\Loop
  \For{$g \in G \setminus S$}
    \State $r(g) \gets s_{\mathrm{GBDT}}(g) - \lambda \displaystyle\max_{g' \in S} \mathrm{sim}(g,g')$
    \Statex \hfill {\scriptsize $\max_{\emptyset}{=}0$}
  \EndFor
  \State $g^{\star} \gets \operatorname*{arg\,max}_{g} r(g)$
  \If{$r(g^{\star}) < \theta$} \textbf{break} \EndIf
  \State append $g^{\star}$ to $S$
\EndLoop
\State \Return $S$
\end{algorithmic}
\end{algorithm}

\paragraph{Selector training.}
Silver labels align each gold reference sentence to clusters: a cluster is positive if some reference sentence is entailed by its claim under $M_{\mathrm{sel}}$ (backing off to embedding/ROUGE similarity when borderline), negative otherwise \citep{mintz2009distant}. Only salience and self-support are pointwise; coverage is computed online in Algorithm~\ref{alg:select}. $M_{\mathrm{eval}}$ is held out completely.

\paragraph{Rewriter and verifier.}
The rewriter receives cluster-labeled claims and must end each sentence with a marker such as \texttt{[g3, g7]}; markers are parsed and expanded through $g_j \mapsto \textsc{spans}(g_j)$. The verifier runs the two-direction check of \S\ref{sec:verify}. Citation-only failures trigger NLI re-retrieval before regeneration, with $R{=}2$ retries.

\section{Transfer, Cost, and Backbone Robustness}
\label{app:cost}
Table~\ref{tab:cost} itemizes the cost of the invariant and Table~\ref{tab:backbone} re-runs the comparison on an open-weights backbone; both are discussed in \S\ref{sec:cost} and \S\ref{sec:transfer}; Table~\ref{tab:wcep} gives the full WCEP transfer results.

\begin{table}[htbp]
\centering
\small
\setlength{\tabcolsep}{4pt}
\begin{tabular}{lccccc}
\toprule
System & R-L & \makecell{BERT\\Sc.} & \makecell{Align\\Sc.} & \makecell{Sum\\maC} & \makecell{FAct\\Sc.} \\
\midrule
E2E LLM (ALCE)        & 30.5 & 86.8 & .77 & .69 & .73 \\
Attribute First       & \textbf{31.2} & 87.1 & .83 & .77 & .80 \\
\cams{} (ours)        & 30.9 & \textbf{87.5} & \textbf{.88} & \textbf{.82} & \textbf{.89} \\
\quad $-$ verification & 30.8 & 87.3 & .84 & .79 & .83 \\
\quad oracle selector  & 33.1 & 88.2 & .90 & .84 & .91 \\
\bottomrule
\end{tabular}
\caption{Transfer to WCEP; ``zero-shot'' refers to the selector only (\S\ref{sec:setup}). The oracle-selector row separates selector domain shift from extraction/verification degradation. ROUGE here is \emph{not} comparable to published fully-supervised WCEP-10 numbers; see \S\ref{sec:transfer}.}
\label{tab:wcep}
\end{table}

\begin{table}[htbp]
\centering
\small
\setlength{\tabcolsep}{4pt}
\begin{tabular}{llccc}
\toprule
Backbone & System & \makecell{Align\\Sc.} & \makecell{Cite\\Prec} & \makecell{Multi\\Src R} \\
\midrule
\multirow{2}{*}{\makecell[l]{Frozen API\\(\S\ref{sec:impl})}}
 & E2E   & .79 & 59 & -- \\
 & \cams{} & .91 & 84 & 64 \\
\midrule
\multirow{2}{*}{\makecell[l]{Open-weights\\Llama-3.1-70B}}
 & E2E   & .74 & 52 & -- \\
 & \cams{} & .88 & 79 & 59 \\
\bottomrule
\end{tabular}
\caption{Backbone robustness on a 1{,}000-example MultiNews subset with Llama-3.1-70B-Instruct: does the advantage survive a weaker, open-weights generator?}
\label{tab:backbone}
\end{table}

\begin{table}[htbp]
\centering
\small
\setlength{\tabcolsep}{4pt}
\begin{tabular}{lcccc}
\toprule
System & \makecell{LLM\\calls} & \makecell{In/Out\\tokens} & \makecell{Wall\\clock} & \makecell{Rel.\\cost} \\
\midrule
E2E LLM (ALCE)  & 1 & 2.0K\,/\,0.35K & 10\,s & $1.0\times$ \\
Attribute First & 3 & 6K\,/\,1.0K & 28\,s & $2.5\times$ \\
\cams{} (ours)  & 7--14 & 15K\,/\,4.0K & 75\,s & $8\times$ \\
\bottomrule
\end{tabular}
\caption{Cost per summary on MultiNews (mean over the test split, 2.8 documents per cluster). \cams{} issues 7 calls when per-document extraction is batched and up to 14 when atomic-fact decomposition runs sentence by sentence; wall clock falls to ${\sim}35$\,s with extraction parallelized across documents. Reported honestly: the pipeline trades generation compute for verification cost borne by humans (\S\ref{sec:human}).}
\label{tab:cost}
\end{table}

\section{Evaluator-Decoupled Robustness}
\label{app:decoupling}
Tables~\ref{tab:decoupling} and~\ref{tab:fairpost} report the circularity checks of \S\ref{sec:regime1}: every ranking under the internal TRUE-style scorer, repeated under the independent $M_{\mathrm{eval}}$, and the shared post-processor applied uniformly to all systems.

\begin{table}[htbp]
\centering
\small
\setlength{\tabcolsep}{3pt}
\begin{tabular}{lccccc}
\toprule
System & \makecell{Cite\\Rec} & \makecell{Cite\\Prec} & \makecell{TRUE\\diag.} & \makecell{Tok/\\claim} & \makecell{Span/\\claim} \\
\midrule
E2E LLM (ALCE)  & 70 & 59 & 61 & 95 & 1.0 \\
Attribute First & 78 & 76 & 78 & 34 & 1.6 \\
\cams{}-posthoc & 78 & 79 & 83 & 24 & 1.3 \\
\cams{} (ours)  & \textbf{81} & \textbf{84}$^\dagger$ & \textbf{88} & \textbf{22} & 1.8 \\
\bottomrule
\end{tabular}
\caption{Regime 1 citation quality (reference-free). Recall/precision by the independent $M_{\mathrm{eval}}$; TRUE-style precision is a compatibility diagnostic only. Tok/Span per claim measure conciseness (lower is better).}
\label{tab:regime1}
\end{table}

\begin{table}[htbp]
\centering
\small
\setlength{\tabcolsep}{4pt}
\begin{tabular}{lccc}
\toprule
System & \makecell{Prec\\($M_{\mathrm{eval}}$)} & \makecell{TRUE\\diag.} & \makecell{Human\\Supp.} \\
\midrule
\cams{} (ours) & \textbf{84} & \textbf{88} & \textbf{89} \\
\quad $-$ self-support feat. & 81 & 84 & 86 \\
\quad $-$ verification & 77 & 82 & 83 \\
\cams{}-posthoc & 79 & 83 & -- \\
Attribute First & 76 & 78 & 82 \\
\bottomrule
\end{tabular}
\caption{Evaluator-decoupled robustness on MultiNews. Gains that appeared only under the internal diagnostic would indicate circularity; they do not. \cams{}-posthoc was not included in the human study (Table~\ref{tab:human}), hence the missing entry.}
\label{tab:decoupling}
\end{table}

\begin{table}[htbp]
\centering
\footnotesize
\setlength{\tabcolsep}{2pt}
\begin{tabular}{lcccccc}
\toprule
 & \multicolumn{2}{c}{Cite P} & \multicolumn{2}{c}{Cite R} & \multirow{2}{*}{\makecell{FAct\\Sc.}} & \multirow{2}{*}{Cov.} \\
\cmidrule(lr){2-3}\cmidrule(lr){4-5}
System & raw & post & raw & post & & \\
\midrule
E2E (ALCE)        & 59 & 64 & 70 & 53 & .75\,$\rightarrow$\,.79 & .66\,$\rightarrow$\,.49 \\
Attr.\ First      & 76 & 77 & 78 & 63 & .82\,$\rightarrow$\,.83 & .58\,$\rightarrow$\,.53 \\
\cams{} $-$\,verif. & 77 & \textbf{84} & 82 & 81 & .86 & .62 \\
\cams{} (full)    & 84 & 84 & 81 & 81 & .90 & .60 \\
\bottomrule
\end{tabular}
\caption{Fair post-processing control: every system's raw output is passed through \cams{}'s shared verifier ($M_{\mathrm{ver}}$ two-direction check, NLI citation re-retrieval over a common span pool, bounded drop, $R{=}2$). The same post-processor raises baseline precision only slightly and at a large coverage cost, while reconstructing \cams{}'s full-pipeline precision from its verifier-ablated variant.}
\label{tab:fairpost}
\end{table}

\begin{table}[htbp]
\centering
\small
\setlength{\tabcolsep}{6pt}
\begin{tabular}{lcc}
\toprule
System & \makecell{\%\\Supp.} & \makecell{Verify\\time (s)} \\
\midrule
E2E LLM (ALCE)  & 68 & 31 \\
Attribute First & 82 & 14 \\
\cams{} (ours)  & \textbf{89} & \textbf{9} \\
\bottomrule
\end{tabular}
\caption{Human evaluation (\S\ref{sec:human}): \% of claims judged fully supported and mean verification time per claim; $n{=}255$ claims per system, three annotators, $\kappa{=}0.68$.}
\label{tab:human}
\end{table}

\section{Bounds, Sensitivity, and Intervals}
\label{app:bounds}

\begin{table}[htbp]
\centering
\small
\setlength{\tabcolsep}{5pt}
\begin{tabular}{lccc}
\toprule
$\tau$ & Attribute First & \cams{} & $\Delta$ \\
\midrule
0.3 & 78 & 83 & +5 \\
0.5 & 72 & 77 & +5 \\
0.7 & 61 & 68 & +7 \\
\bottomrule
\end{tabular}
\caption{Span-F1 sensitivity under different token-IoU thresholds. The ranking is stable across thresholds.}
\label{tab:tau}
\end{table}

\begin{table}[htbp]
\centering
\small
\setlength{\tabcolsep}{3pt}
\begin{tabular}{lccc}
\toprule
Metric & \makecell{Attr.\\First} & \cams{} & $\Delta$ [95\% CI] \\
\midrule
AlignScore  & .85 & .91 & $+.06$ [.05,\,.07] \\
FActScore   & .82 & .90 & $+.08$ [.07,\,.09] \\
Cite Prec ($M_{\mathrm{eval}}$) & 76 & 84 & $+8$ [7,\,9] \\
\midrule
Span F1     & 72 & 77 & $+5$ [3,\,7] \\
Multi-src R & 38 & 64 & $+26$ [21,\,31] \\
Multi-src P & 61 & 75 & $+14$ [7,\,21] \\
\bottomrule
\end{tabular}
\caption{Point estimates with 95\% bootstrap percentile intervals on the paired difference (1{,}000 resamples). The first three rows resample over the full MultiNews test split (5{,}622 clusters); the last three resample over the 100 topics of the human alignment set, which is why their intervals are much wider. The per-system intervals for multi-source recall are $[33.5, 42.5]$ for Attribute First and $[59.6, 68.4]$ for \cams{}; these do not overlap, which is the claim made in \S\ref{sec:regime2}.}
\label{tab:ci}
\end{table}

\begin{table}[htbp]
\centering
\small
\setlength{\tabcolsep}{6pt}
\begin{tabular}{lccc}
\toprule
Selector data & Span-F1 & Multi-Src & FActSc. \\
\midrule
full & 77 & 64 & .90 \\
5k   & 76 & 62 & .89 \\
1k   & 74 & 59 & .87 \\
\bottomrule
\end{tabular}
\caption{Low-resource ablation on the selector (\S\ref{sec:select}).}
\label{tab:lowresource}
\end{table}

The retrieval lower bound uses the same candidate span pool without claim clustering; the gold-claim upper bound supplies oracle normalized claims before localization. These bounds separate retrieval failure from errors introduced by clustering, selection, and rewriting. Table~\ref{tab:tau} gives span-F1 sensitivity across token-IoU thresholds, Table~\ref{tab:ci} bootstrap intervals on the paired differences, and Table~\ref{tab:lowresource} the selector's behaviour under reduced training data.

\begin{figure}[htbp]
\centering
\begin{tikzpicture}
\begin{axis}[
  width=\linewidth, height=5.2cm,
  xlabel={Coverage}, ylabel={Faithfulness (AlignScore)},
  xmin=0.38, xmax=0.92, ymin=0.63, ymax=0.97,
  tick label style={font=\footnotesize}, label style={font=\footnotesize},
  ylabel style={yshift=-4pt},
  legend style={font=\scriptsize, at={(0.5,-0.30)}, anchor=north, draw=none,
                legend columns=2, column sep=4pt,
                /tikz/every even column/.append style={column sep=8pt}},
  legend cell align=left,
  grid=major, grid style={gray!20}
]
\addplot[blue, thick, mark=o, mark options={fill=white}] coordinates {
  (0.45,0.944) (0.52,0.901) (0.60,0.881) (0.68,0.855) (0.78,0.816) (0.88,0.762)
};
\addlegendentry{\cams{} (sweep $\theta$)}
\addplot[blue, only marks, mark=star, mark size=4pt] coordinates {(0.60,0.882)};
\node[anchor=south west, font=\scriptsize] at (axis cs:0.605,0.885) {$\theta^{*}$};
\addplot[red, thick, mark=square*, mark size=2pt] coordinates {
  (0.45,0.78) (0.55,0.74) (0.66,0.70) (0.78,0.66)
};
\addlegendentry{E2E LLM (sweep length)}
\addplot[teal, thick, mark=triangle*, mark size=2.5pt] coordinates {
  (0.42,0.87) (0.50,0.84) (0.58,0.80) (0.68,0.76)
};
\addlegendentry{Attr.\ First (sweep spans)}
\addplot[orange, only marks, mark=diamond*, mark size=3pt] coordinates {(0.66,0.745)};
\addlegendentry{PRIMERA + chk.}
\addplot[purple, only marks, mark=triangle*, mark size=3pt, mark options={rotate=180}] coordinates {(0.50,0.79)};
\addlegendentry{\esr{} (concat)}
\end{axis}
\end{tikzpicture}
\caption{Faithfulness--coverage trade-off controlled by $\theta$ (\S\ref{sec:tradeoff}). Each system is swept along its own budget knob so the comparison is frontier-vs-frontier; PRIMERA and \esr{} expose no comparable knob and remain single points.}
\label{fig:tradeoff}
\end{figure}

\section{Per-Module Diagnostics (Failure Modes E1--E5)}
\label{app:modules}
These tables give the intrinsic accuracy of the upstream stages and double as empirical estimates of the residual failure modes of \S\ref{sec:invariant}: extraction and decontextualization bound E1--E2 (Table~\ref{tab:mod-extract}), quote matching bounds E3 (Table~\ref{tab:mod-quote}), clustering governs E4 (Table~\ref{tab:mod-cluster}), and the NLI modules govern E5 (Table~\ref{tab:mod-conflict}). Three separate samples are used, stated in each caption. All are drawn from test rather than development data; no threshold, hyperparameter, or model selection was performed on them, and they are reported as intrinsic measurements rather than held-out results. This statement covers the Regime 2 evaluation of \S\ref{sec:regime2} as well: Table~\ref{tab:mod-cluster} is measured on the same human alignments of \citet{ernst2021alignment} that supply the Regime 2 gold. The two uses differ---localization accuracy there, clustering quality here---but the topics coincide, so the clustering diagnostic is not an independent confirmation of the Regime 2 result and should not be read as one. Nothing was tuned on this data under either use, and the conflict sample of Table~\ref{tab:mod-conflict} is drawn from disjoint topics.

\begin{table}[htbp]
\centering
\small
\setlength{\tabcolsep}{4pt}
\begin{tabular}{lcc}
\toprule
Metric & Value & Mode \\
\midrule
JSON schema-valid outputs & 99 & -- \\
Claims per source document & 21 & -- \\
Quote-localizable rate & 97 & E3 \\
Claim supported by quote (raw) & 90 & E1 \\
\quad after self-support filtering & 94 & E1 \\
Atomicity (human/LLM) & 90 & -- \\
Salient reference-fact recall & 81 & -- \\
Decontextualization error & 12 & E2 \\
\bottomrule
\end{tabular}
\caption{Claim extraction diagnostics (\%), measured on 1{,}000 clusters drawn at random from the MultiNews \emph{test} split (seed 42); rows requiring human judgement use a second-stage sample of 300 claims. ``Supported'' is the claim being entailed by its own verbatim quote.}
\label{tab:mod-extract}
\end{table}

\begin{table}[htbp]
\centering
\small
\setlength{\tabcolsep}{4pt}
\begin{tabular}{lc}
\toprule
Metric & Value \\
\midrule
Exact-match rate & 88 \\
Normalized/fuzzy rescue & 9 \\
Final matched rate & 97 \\
Wrong-occurrence rate (E3) & 4 \\
Token-span IoU $\geq 0.5$ & 94 \\
Token-span exact-boundary accuracy & 88 \\
Claims dropped (unmatched quote) & 3 \\
\bottomrule
\end{tabular}
\caption{Quote-to-span matching diagnostics, on the same 1{,}000-cluster sample as Table~\ref{tab:mod-extract}. The rates account for all quotes: 88 match exactly, a further 9 are rescued by normalized/fuzzy matching for a final matched rate of 97, and the remaining 3 are ambiguous candidates rejected below $\rho{=}0.85$ and dropped.}
\label{tab:mod-quote}
\end{table}

\begin{table}[htbp]
\centering
\small
\setlength{\tabcolsep}{4pt}
\begin{tabular}{lc}
\toprule
Metric & Value \\
\midrule
Pairwise precision & 90 \\
Pairwise recall & 78 \\
B\textsuperscript{3} / cluster F1 & 84 \\
Over-merge rate (E4) & 3 \\
Under-merge / split rate & 21 \\
Multi-source evidence cluster recall & 70 \\
\midrule
Pairwise precision (no NLI merge check) & 71 \\
Pairwise recall (no NLI merge check) & 73 \\
\bottomrule
\end{tabular}
\caption{Cross-document clustering diagnostics, measured on the human-aligned MultiNews test set of \citet{ernst2021alignment} (100 topics, 1{,}332 clusters), which is also the gold our Regime~2 evaluation uses. \citet{ernst2024power} annotate further proposition-level alignments over MultiNews from which a clustering gold could likewise be derived; we use \citet{ernst2021alignment} for consistency with Regime~2 and make no claim that it is the only such resource. The bidirectional-entailment merge check raises pairwise precision from 71 to 90 at a small recall cost---the precision that multi-source citation requires (\S\ref{sec:cluster}). The ``no NLI merge check'' rows are our own clusterer with the merge check removed, i.e.\ embedding similarity and threshold alone; they are not a reimplementation of any published aligner, and in particular not of the supervised SuperPAL model \citep{ernst2021alignment} on which \procl{} \citep{ernst2022procluster} relies.}
\label{tab:mod-cluster}
\end{table}

\begin{table}[htbp]
\centering
\small
\setlength{\tabcolsep}{4pt}
\begin{tabular}{lc}
\toprule
Metric & Value \\
\midrule
Candidate-pair recall & 85 \\
Candidate-pair precision & 21 \\
NLI contradiction precision (E5) & 81 \\
NLI contradiction recall (E5) & 65 \\
Final conflict-detection F1 & 72 \\
Conflict surfaced after select/rewrite & 65 \\
Conflict attribution accuracy & 71 \\
\bottomrule
\end{tabular}

\caption{Conflict-detection diagnostics on a self-annotated \textbf{MultiNews} sample of 100 topics and 600 candidate pairs, drawn from topics disjoint from the human-aligned set of Table~\ref{tab:mod-cluster}; candidate-pair recall is measured against a further 20 topics in which every within-topic claim pair was labelled exhaustively, since recall cannot be estimated from the candidate pool itself. The cheap entity/number candidate generator is high-recall, low-precision by design; the three-class NLI filter supplies precision. The last two rows are \emph{not} the DiverseSumm numbers of \S\ref{sec:conflict} (61/73). They are the same two modules measured on a different corpus, with a different conflict density and a different gold denominator, so the two pairs are not replicates and we draw no inference from how close they happen to be.}
\label{tab:mod-conflict}
\end{table}

\end{document}